\documentclass{article}  
\usepackage{iclr2027_conference,times}   
\usepackage{subcaption} 
\usepackage{hyperref}            
\usepackage{url}

\usepackage{tabularx}
\usepackage{times}
\usepackage{latexsym}
\usepackage[T1]{fontenc}
\usepackage[utf8]{inputenc}
\usepackage{microtype}
\usepackage{inconsolata}
\usepackage{graphicx}
\usepackage{comment}
\usepackage[most]{tcolorbox}

\tcbuselibrary{listings,skins}
\usepackage{listings}

\definecolor{vscodekeyword}{RGB}{175,0,120}
\definecolor{vscodefunc}{RGB}{70,70,210}
\definecolor{vscodenum}{RGB}{210,90,40}
\definecolor{vscodestring}{RGB}{40,130,40}
\definecolor{vscodecomment}{RGB}{120,120,120}

\lstdefinestyle{stabpython}{
    language=Python,
    basicstyle=\ttfamily\footnotesize,
    keywordstyle=\bfseries\color{vscodekeyword},
    emphstyle=\bfseries\color{vscodefunc},
    commentstyle=\itshape\color{vscodecomment},
    stringstyle=\color{vscodestring},
    emph={len,range,list,str,map,join,extend},
    numbers=left,
    numberstyle=\scriptsize\color{gray!60},
    numbersep=6pt,
    xleftmargin=1.8em,
    frame=none,
    breaklines=true,
    columns=fullflexible,
    keepspaces=true,
    showstringspaces=false,
    aboveskip=0pt,
    belowskip=0pt,
    literate=
        *{0}{{{\color{vscodenum}0}}}{1}
         {1}{{{\color{vscodenum}1}}}{1}
         {2}{{{\color{vscodenum}2}}}{1}
         {3}{{{\color{vscodenum}3}}}{1}
         {4}{{{\color{vscodenum}4}}}{1}
         {5}{{{\color{vscodenum}5}}}{1}
         {6}{{{\color{vscodenum}6}}}{1}
         {7}{{{\color{vscodenum}7}}}{1}
         {8}{{{\color{vscodenum}8}}}{1}
         {9}{{{\color{vscodenum}9}}}{1}
}

\newtcblisting{stabcode}[2][]{
    listing only,
    width=\linewidth,
    colback=blue!7,
    colframe=blue!65!black,
    coltitle=white,
    colbacktitle=blue!55!black,
    title={#2},
    fonttitle=\bfseries\small,
    boxrule=0.7pt,
    arc=1.2mm,
    left=1.5mm,
    right=1.5mm,
    top=1mm,
    bottom=1mm,
    enhanced,
    listing options={style=stabpython},
    #1
}

\usepackage{amsfonts}
\usepackage{amsmath}
\usepackage{amssymb}
\usepackage{amsthm}
\usepackage{booktabs}
\usepackage{cleveref}
\usepackage{enumitem}
\usepackage{multirow}
\usepackage{soul}
\usepackage{xcolor}

\usepackage{arydshln}

\title{STAB: Specification-driven Testing for Algorithmic Bottlenecks}

\author{
Soohan Lim\textsuperscript{*},
Joonghyuk Hahn\textsuperscript{*},
Hyundong Jin,
Yo-Sub Han\textsuperscript{$\dagger$}\\
Yonsei University, Seoul, Republic of Korea\\
   \texttt{\{%
   \href{mailto:aness1219@yonsei.ac.kr}{aness1219},%
   \href{mailto:greghahn@yonsei.ac.kr}{greghahn},%
   \href{mailto:tuzi04@yonsei.ac.kr}{tuzi04},%
   \href{mailto:emmous@yonsei.ac.kr}{emmous}%
   \}@yonsei.ac.kr}
}

\iclrfinalcopy

\begin{document}
\maketitle

\lhead{Preprint. Under review.}

\begingroup
\renewcommand\thefootnote{\fnsymbol{footnote}}
\footnotetext[1]{Equal contribution.}
\footnotetext[2]{Corresponding author.}
\endgroup

\begin{abstract}
Evaluating the efficiency of algorithmic code requires test cases that expose runtime bottlenecks.
Previous methods generate efficiency test cases either by increasing input size or 
by generating code-specific inputs that make the given implementation run slowly. 
Consequently, they do not address the structural input conditions that drive the algorithmic worst case.
We introduce \textbf{STAB}, a specification-driven pipeline that generates test cases that expose algorithmic bottlenecks 
from a natural-language problem specification alone.
STAB separates the task into constraint-bound maximization and adversarial structure injection.
(i) The constraint saturator extracts constraints and resolves large admissible size assignments using rule-based saturation and CP-SAT optimization over related variables.
(ii) The adversarial scenario injector retrieves implementation-level adversarial construction principles 
from a curated scenario catalog using keyword matching and K-nearest neighbors~(KNN).
STAB encodes the problem specification, resolved boundary, and retrieved construction principles into a structured generation specification, 
from which the LLM synthesizes a Python generator function.
On CodeContests, STAB raises the rate of generated test cases that expose algorithmic bottlenecks from 50.43\% to 73.45\% on
average across open-source LLMs and from 57.45\% to 71.85\% on average across closed-source LLMs,
with consistent gains across Python, Java, and C++. 
Our code is available at \url{https://github.com/suhanmen/STAB}.

\end{abstract}

\section{Introduction}\label{sec:intro}
Large language models~(LLMs) are widely used for code generation~\citep{chen2021codex,abs-2401-14196,abs-2409-12186} and 
test generation~\citep{yuan2024chatgpt,chen2024chatunittest,ryan2024codeaware,schafer2024empirical,lim2026contracteval}.
Existing benchmarks and test generation methods mainly check whether a program returns the expected output on a fixed test suite.
For algorithmic problems with time limits, however, passing ordinary tests does not guarantee efficiency on large inputs allowed by the specification.
This setting requires efficiency test cases, valid inputs that expose algorithmic bottlenecks rather than output correctness.

Generating efficiency test cases is not equivalent to making inputs large.
Worst-case behavior often depends on structure as well as size.
Reverse-sorted arrays stress pivot-sensitive sorting, path-shaped trees increase traversal depth, 
and dense graphs increase the work of shortest path routines.
An efficiency test case must therefore combine a large valid input boundary with a structure that increases runtime.
Existing efficiency-oriented test generators address only part of this requirement.
EvalPerf~\citep{evalperf} scales an LLM-generated input until a target implementation reaches a runtime or memory threshold.
WEDGE~\citep{wedge} infers bottleneck conditions from executions of a target program and searches for inputs satisfying those conditions.
Both approaches rely on input scaling or feedback from a specific implementation.
They are useful for slowing a given program, but less suited to generating tests from the problem specification alone.

We propose \textbf{STAB}~(\textbf{S}pecification-driven \textbf{T}esting for \textbf{A}lgorithmic \textbf{B}ottlenecks), 
a pipeline that generates efficiency test cases from a natural-language problem specification alone.
STAB separates the task into constraint-bound maximization and adversarial structure injection.
(i) The constraint saturator extracts input constraints from the specification and resolves 
the largest size parameters those constraints allow by considering relationships among variables 
through rule-based saturation and CP-SAT~\citep{ortools} optimization.
(ii) The adversarial scenario injector matches the problem to scenarios in a curated catalog through keyword matching and KNN,
and retrieves the associated construction principles.
STAB then encodes the problem specification, resolved boundary, and retrieved construction principles into a structured generation specification, 
from which the LLM synthesizes a Python generator function.
This decomposition keeps the LLM's role narrow:
it composes a valid generator from a resolved boundary and selected construction principles,
rather than solving constraint satisfaction and adversarial input design at once.
We evaluate STAB on CodeContests~\citep{abs-2203-07814} across Python, Java, and C++, using four LLMs 
that cover open-source and closed-source settings.
For each generated test case, we count success when the test case drives an accepted solution to a runtime larger than the maximum runtime induced 
by the CodeContests test set for the same problem.
STAB increases the share of generated test cases that expose algorithmic bottlenecks from 50.43\% to 73.45\% on average across open-source LLMs and
from 57.45\% to 71.85\% on average across closed-source LLMs.

We make three main contributions:
\begin{itemize}[leftmargin=*]
    \item \textbf{Constraint-aware boundary resolution.}
    We introduce a constraint saturator that extracts constraints from specifications and computes large valid size assignments over related variables.
    \item \textbf{Scenario-guided adversarial construction.}
    We build a curated catalog of adversarial construction principles and retrieve problem-matched principles using keyword matching and KNN.
    \item \textbf{Empirical validation across models and languages.}
    We show that STAB outperforms standard prompting and prior efficiency-oriented generators across the CodeContests evaluation.
\end{itemize}

\section{Related Work}\label{sec:related}

\subsection{Efficiency in LLM-based Code Generation}\label{ssec:related-codegen}
Research on LLM-based code generation has centered on functional correctness.
Benchmarks such as HumanEval, MBPP, and APPS measure whether a generated program passes a fixed reference test 
suite~\citep{chen2021codex, austin2021mbpp, hendrycks2021apps}. 
Code-specialized models such as CodeLlama, DeepSeek-Coder, StarCoder, and the 
Qwen-Coder series have steadily raised this correctness 
baseline~\citep{roziere2023codellama, abs-2401-14196, li2023starcoder, abs-2409-12186}. 
AlphaCode and its follow-ups have extended LLM-based code generation to competitive programming,
where problems typically specify input constraints and time limits and require algorithmic reasoning.
In this setting, the gap between a functionally correct solution and 
an efficient solution becomes pronounced~\citep{abs-2203-07814, ridnik2024alphacodium}.
Prior work has also studied the efficiency of generated code.
EffiLearner uses execution overhead profiles to improve the efficiency of LLM-generated code, 
and PerfCodeGen incorporates runtime feedback during self-refinement~\citep{0005DWWQCG024,peng2025perfcodegen}.
These studies show that functional correctness alone does not fully characterize the quality of generated code.
For algorithmic code generation, these results indicate that efficiency evaluation must accompany correctness evaluation.

\subsection{LLM-assisted Test Case Generation}\label{ssec:related-tc}
LLM-assisted test case generation has developed along multiple directions.
Functional validation methods synthesize unit tests that check whether a program returns the expected output, 
with representative work including EvalPlus, ChatUniTest, and CodaMOSA~\citep{liu2023evalplus, chen2024chatunittest, lemieux2023codamosa}.
Robustness-oriented methods generate failure-inducing inputs that expose correctness or robustness failures in candidate programs~\citep{wang2023recode, zhang2025exlong, zhong2025exlongtool, lim2026contracteval}.
Efficiency evaluation methods generate test inputs that stress runtime behavior to assess the efficiency of candidate programs.
EvalPerf~\citep{evalperf} synthesizes an input generator and scales a size parameter until the measured runtime reaches a threshold.
WEDGE~\citep{wedge} identifies constructs at the implementation level,
such as loops, and generates inputs that increase the number of operations those constructs perform.
In both cases, the generation signal comes from input size scaling or execution feedback from a particular implementation. 
Neither method explicitly infers the algorithmic structure implied by the problem or targets the slow input construction 
associated with that structure.
Algorithmic problems, however, require efficiency test inputs that are both valid at large input sizes 
and structured to trigger worst-case behavior.
STAB addresses this gap by generating efficiency test cases from the problem specification alone.
It resolves input constraints and encodes adversarial constructions implied by the algorithmic problem, 
rather than relying on input size or behavior observed from a specific implementation.

\section{Notation}\label{sec:notation}
All notation is defined for a fixed implementation language, and the language
index is omitted because Python, Java, and C++ are evaluated independently.
For an evaluated CodeContests problem \(p\), let
\(\mathcal{T}_{\mathrm{CC}}(p)\) and \(\mathcal{T}_{\mathrm{gen}}(p)\)
denote the CodeContests test set and the generated efficiency test set, respectively.
The set \(\mathcal{A}(p)\) contains the accepted solutions for \(p\), and
\(\tau(s,t)\) denotes the execution time of solution \(s\) on test case \(t\).

For each problem \(p\), we selected five \emph{evaluation solutions} from \(\mathcal{A}(p)\).
The solutions \(s_{p,\mathrm{fast}}\) and \(s_{p,\mathrm{slow}}\) have the
lowest and highest mean runtimes on \(\mathcal{T}_{\mathrm{CC}}(p)\),
respectively, whereas \(s_{p,r_i}\), \(i \in \{1,2,3\}\), were sampled at random.
We use \(s_{p,r}\) to denote any of these five evaluation solutions.
In the main evaluation, the \emph{reference implementation} supplied to EvalPerf or WEDGE was excluded from this set.







\section{The STAB Pipeline}\label{sec:method}
\begin{figure*}[t] 
    \centering
    \includegraphics[width=\textwidth]{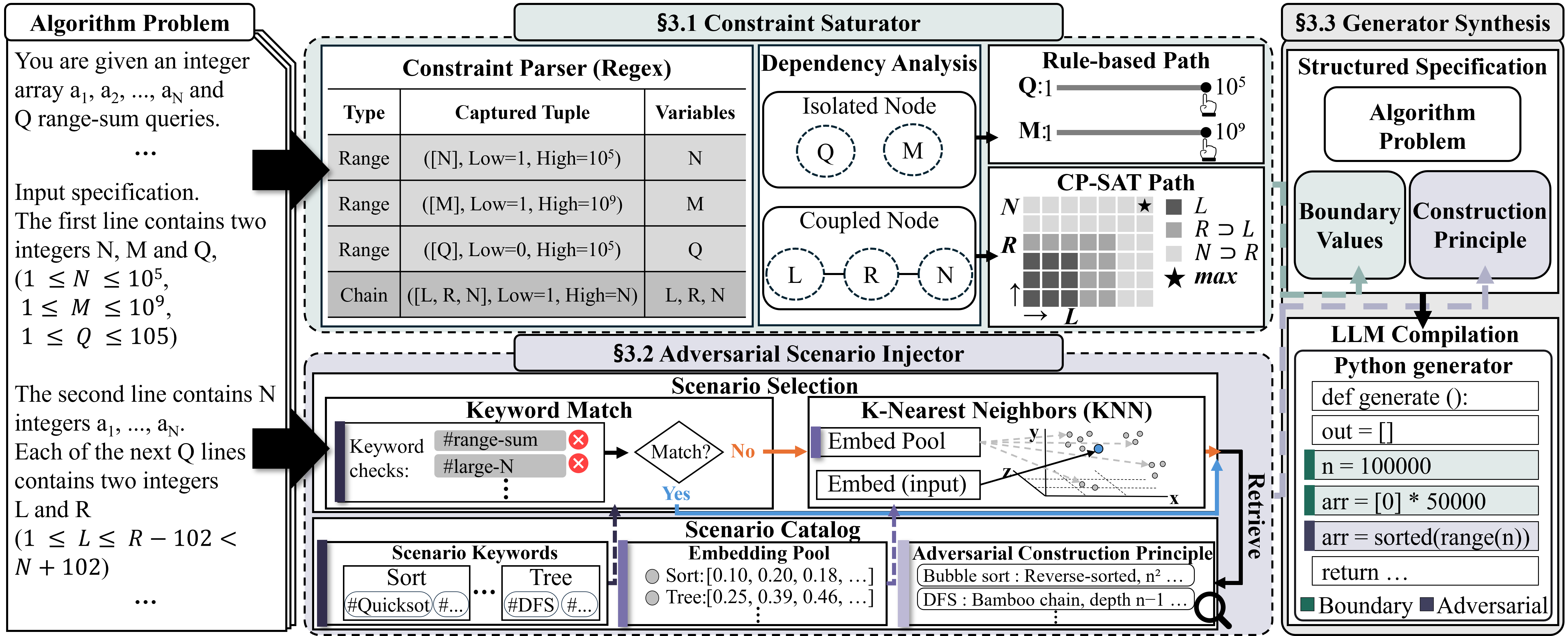}
    \caption{
    An overview of the STAB pipeline.
    }
    \label{fig:Overview}
\end{figure*}

Given the natural-language specification of problem \(p\), STAB constructs
\(\mathcal{T}_{\mathrm{gen}}(p)\) without using a reference implementation.
For each \(t \in \mathcal{T}_{\mathrm{gen}}(p)\), STAB synthesizes a Python
generator function that materializes the input.
A direct LLM prompt must infer the valid boundary and the input construction
from the problem specification while writing the generator.
This can lead the model to satisfy obvious numeric bounds while missing the
adversarial structure or to propose a plausible structure that violates the
input format.
STAB reduces this burden by supplying the final LLM call with two intermediate
outputs.
(i) The constraint saturator computes the largest admissible values for
variables that define the valid input boundary.
(ii) The adversarial scenario injector selects scenarios that match the problem
and retrieves their construction principles.
The LLM then composes a generator from the problem specification, the resolved
boundary, and the selected principles.
Figure~\ref{fig:Overview} shows the process.

\begin{table*}[t]
\centering
\small
\setlength{\tabcolsep}{5pt}
\renewcommand{\arraystretch}{1.1}
\caption{
Example adversarial construction principles from STAB's scenario catalog.
Rows pair representative implementations with construction principles encoded in the structured generation specification.
The full catalog of 13 scenarios and 51 implementations appears in Appendix~\ref{app:scenario_table}.
}
\begin{tabularx}{\textwidth}{
@{}
>{\raggedright\arraybackslash}p{2.0cm}
>{\raggedright\arraybackslash}p{3.6cm}
>{\raggedright\arraybackslash}X
@{}
}
\toprule
\textbf{Scenario} & \textbf{Implementation} & \textbf{Adversarial Construction Principle} \\
\midrule
Sorting
& Quicksort with first or last pivot
& An already sorted sequence makes the pivot an extreme element at each partition, creating highly unbalanced recursion. \\
\addlinespace[0.25em]
Hashing
& Hash table with identity hash
& Keys chosen as multiples of the bucket modulus collide into one bucket, so lookup traverses a long chain. \\
\addlinespace[0.25em]
Tree traversal
& DFS via system recursion
& A bamboo tree with edges \((i, i+1)\) for \(1 \le i < n\) gives depth \(n-1\) and maximizes recursion depth. \\
\bottomrule
\end{tabularx}
\label{tab:algorithm_catalog}
\end{table*}

\subsection{Constraint Saturator}\label{ssec:bound_solver}
The constraint saturator computes the largest admissible values for variables that define the valid input boundary of an algorithmic problem.
Its output is a \emph{resolved boundary}: an assignment to variables such as the number of elements, queries, vertices, or edges 
that satisfies the constraints extracted from the problem specification.
This step accounts for relationships among variables rather than maximizing each variable independently.
Variables that appear only in unary bounds can be saturated to their largest extracted values.
Variables that appear together in product bounds, chained inequalities, or aggregate constraints must be resolved 
jointly because increasing one variable can restrict the valid range of another.
The constraint saturator therefore applies rule-based saturation to independent variables and CP-SAT optimization to coupled variable groups.
The resulting boundary is encoded together with the selected construction principles and the problem specification into the structured generation specification, 
from which the LLM compiles a generator that instantiates the corresponding constructions within a validated input range.

\paragraph{Constraint Extraction.}
Algorithmic problem specifications often express input constraints as mathematical notation embedded
in natural language, such as $1 \le N \le 10^5$.
The constraint saturator first converts these constraints into normalized records that contain variables, relation types, and bounds.
A regular expression parser is suitable here because the same constraint patterns repeatedly appear across problem specifications,
including ranges, chained inequalities, products, and string length bounds.
For example, it maps
$1 \le N \le 10^5$ to a scalar range,
$1 \le a_i \le 10^9$ to an indexed range,
$1 \le L \le R \le N$ to a chained inequality,
$N \cdot M \le 10^6$ to a product constraint,
and $|s| \le 10^5$ to a string length bound.
The parser returns normalized constraint records containing variable names, constraint types,
and integer bounds, which are used in dependency analysis.

\paragraph{Dependency Analysis.}

After constraint extraction, the constraint saturator builds a variable dependency graph.
Each node is a variable in the normalized constraint records.
A constraint relating two or more variables creates dependency edges among them.
Variables with no dependency edge are treated as independent variables.
Variables connected by dependency edges form a dependent variable group.
Each dependent variable group is resolved jointly because increasing one variable can reduce the valid range of another.
For example, $N$ is treated as dependent when it appears not only in $1 \le N \le 10^5$
but also in a product constraint such as $N \cdot M \le 10^6$ or in a chain such as $1 \le L \le R \le N$.
The graph groups variables rather than constraints before boundary resolution.

\paragraph{Boundary Resolution.}
The constraint saturator applies different procedures to independent variables and dependent variable groups.

\begin{itemize}
    \item \textbf{Rule-based Path.}
    For each independent variable, the constraint saturator assigns the largest upper bound extracted for that variable.
    Since the variable has no dependency edge, increasing it to its bound cannot violate any constraint involving another variable.

    \item \textbf{CP-SAT Path.}
    For each dependent variable group, the constraint saturator formulates a CP-SAT model from the variable domains 
    and the extracted constraints that relate variables in the group~\citep{ortools}.
    The CP-SAT solver returns a feasible assignment that makes the variables in the group large while
    respecting the extracted constraints. This path is used for chained inequalities,
    product constraints, and aggregate size constraints, where independent saturation can produce an invalid assignment.
\end{itemize}

The resolved boundary is not a proof of the global maximum over the full input space.
It is a valid size configuration under the constraints extracted by the constraint saturator and 
serves as the validated size budget for generator construction.

\subsection{Adversarial Scenario Injector}\label{ssec:adversarial_scenario_injector}
The adversarial scenario injector matches the problem specification to scenarios
in the catalog using keyword matching and KNN.
It supplies construction principles from the matched scenarios to the generator construction step.
Among the sorting entries, one construction principle specifies a reverse-sorted sequence for pivot-sensitive sorting strategies.
Given the supplied principles and the resolved boundary, the LLM composes a generator that emits the resolved boundary values
for size-determining variables and encodes the input structure according to the selected principles.

\paragraph{Scenario Catalog.}
We maintain a catalog of 13 adversarial scenarios, each corresponding to an algorithm family
such as sorting, hashing, tree traversal, or graph shortest path.
For each scenario, the catalog stores 
(i) representative implementations and construction principles that trigger their worst-case behavior,
(ii) scenario keywords used by the keyword matching step, and (iii) an anchor pool for KNN retrieval.
We construct the anchor pool by matching unlabeled training problem descriptions to each scenario through its keywords.
The descriptions come from a training split independent of our evaluation set, and we do not access reference implementations,
generated tests, runtime measurements, or evaluation outcomes.
The catalog is compiled from standard algorithms and competitive programming references~\citep{sedgewick2011,laaksonen2017} 
and foundational work on algorithmic complexity attacks~\citep{mcilroy1999killer,crosby2003}.
Table~\ref{tab:algorithm_catalog} summarizes representative adversarial construction principles from the scenario catalog.

\paragraph{Scenario Selection.}
The adversarial scenario injector retrieves candidate scenarios through two
mutually exclusive paths: Keyword Match and KNN.
Keyword Match retrieves every scenario whose keywords appear explicitly in
the problem specification.
KNN is invoked only when Keyword Match retrieves no scenario.
It returns the two scenarios with the highest cosine similarity.
The outputs of the two paths are never merged or jointly ranked.
This conditional rule prioritizes explicit lexical evidence when available
and uses semantic similarity only as a fallback, reducing irrelevant
construction principles passed to generator synthesis.
Appendix~\ref{app:retrieval-ablation} compares this rule with seven
alternative retrieval configurations.

\begin{itemize}
    \item \textbf{Keyword Match.}
        Scenario keywords take two forms.
        The first consists of algorithm or data structure names, such as ``Dijkstra'' for graph shortest path.
        The second consists of task-level cues, such as ``non-decreasing'' for sorting or ``range query'' for BST and segment structures.
        At inference, Keyword Match scans the problem specification and returns every scenario whose keywords appear.
    
    \item \textbf{K-Nearest Neighbors~(KNN).}
        For each scenario, we embed its anchors with \texttt{SFR-Embedding-2\_R}, L2-normalize 
        the embeddings, and average them into a single scenario centroid.
        At inference, we embed the evaluation problem with the same model,
        compare it to the centroids by cosine similarity, and retain the top two scenarios.

\end{itemize}

\subsection{Generator Synthesis}\label{ssec:generator_synthesis}

STAB encodes the generation task as a structured 
specification over three components: 
the resolved boundary from the constraint saturator,
the construction principles from the adversarial scenario injector,
and the problem specification.
The LLM compiles this specification into a Python generator.


\section{Experimental Setup}\label{sec:setting}

\subsection{Datasets}\label{ssec:dataset-construction}
We used the validation and test splits of CodeContests~\citep{abs-2203-07814}
for runtime evaluation in Python, Java, and C++.
The training split was used only to build the KNN anchor pool from
problem specifications~(\S\ref{ssec:adversarial_scenario_injector}),
without solutions, generated tests, runtime measurements, or evaluation outcomes.
For each problem \(p\) and implementation language, we ranked all
\(s \in \mathcal{A}(p)\) by mean runtime on \(\mathcal{T}_{\mathrm{CC}}(p)\)
and selected the fastest, slowest, and three randomly sampled
evaluation solutions.
Appendix~\ref{app:impl_detail} reports problem counts;
Appendix~\ref{app:dataset} details dataset construction and solution selection.

\subsection{Evaluated Models and Baselines}\label{sec:Baselines}

\paragraph{Evaluated Models.}

We evaluate four LLMs: two open-source instruction-tuned models with comparable 
parameter counts and publicly accessible weights, Qwen-3.5-27B~(Qwen-3.5) and 
gemma-4-31B-it~(Gemma-4), and two closed-source models, OpenAI's gpt-5.4-nano~(GPT-5.4) 
and Google's gemini-3.1-flash-lite~(Gemini-3.1), with undisclosed parameter counts 
and training data.

\paragraph{Baseline Strategies.}
The evaluation compares STAB against three baselines under two input settings.
Standard Prompting~(Base) uses only the algorithmic problem specification as input, 
matching STAB's input condition, and generates a Python generator function that emits an efficiency test case.
EvalPerf~\citep{evalperf} and WEDGE~\citep{wedge} additionally require a reference implementation
and therefore operate under richer input than STAB.
We include them as representative prior methods for efficiency test generation.

\subsection{Evaluation Metric}
We define Algorithmic Slowdown Rate~(ASR) by comparing the runtime induced by
each generated efficiency test case \(t \in \mathcal{T}_{\mathrm{gen}}(p)\)
with the largest runtime induced by \(\mathcal{T}_{\mathrm{CC}}(p)\) on the
same evaluation solution \(s_{p,r}\).
Runtime \(\tau(s,t)\) is measured in the
DOMjudge~\citep{domjudge} environment described in
Appendix~\ref{app:domjudge}.
ASR is computed separately for each implementation language.

Let \(\mathcal{P}\) be the set of evaluated problems.
For each problem \(p\) and evaluation solution \(s_{p,r}\), the
CodeContests runtime threshold is
\[
\tau_{p,r}^{*}
=
\max_{t\in\mathcal{T}_{\mathrm{CC}}(p)}
\tau(s_{p,r},t).
\]

With
\(N_{\mathrm{gen}}
=\sum_{p\in\mathcal{P}}
|\mathcal{T}_{\mathrm{gen}}(p)|\),
ASR for a fixed evaluation solution index \(r\) is
\[
\mathrm{ASR}_{r}
=
\frac{1}{N_{\mathrm{gen}}}
\sum_{p\in\mathcal{P}}
\sum_{t\in\mathcal{T}_{\mathrm{gen}}(p)}
\mathbb{I}\!\bigl[
\tau(s_{p,r},t)>\tau_{p,r}^{*}
\bigr],
\]
where \(\mathbb{I}[\cdot]\) returns 1 when the generated test case exposes
a bottleneck and 0 otherwise.

\section{Results and Analysis}\label{sec:result_analysis}
We first compare STAB against Base prompting on the CodeContests benchmark in Section~\ref{ssec:comparison-with-baselines}.
The remaining sections address two questions through complementary analyses.
Sections~\ref{ssec:reference-variance} and~\ref{ssec:module-ablation}
examine whether STAB produces stable gains with a clear module-level
decomposition, focusing on variation across heterogeneous evaluation
solutions and the contribution of each pipeline module.
Sections~\ref{ssec:prior-comparison} and~\ref{ssec:worst-case-comparison}
then place STAB in context by comparing it against prior efficiency-oriented
generators and the CodeContests test set
\(\mathcal{T}_{\mathrm{CC}}(p)\).


\subsection{Main Results}\label{ssec:comparison-with-baselines}
Table~\ref{tab:full_result} reports ASR, the fraction of tests
\(t \in \mathcal{T}_{\mathrm{gen}}(p)\) for which
\(\tau(s_{p,r},t) > \tau_{p,r}^{*}\).
STAB improves the average for all four LLMs, raising the model average from 53.94\% to 72.65\%.
At the model level, the relative gains from Base to STAB are 31.56\% for Qwen-3.5, 61.90\% for Gemma-4, 49.26\% for Gemini-3.1, and 7.20\% for GPT-5.4. 
Appendix~\ref{app:execution-validity} reports execution validity.

The \textit{Fast} and \textit{Slow} groups contain
\(s_{p,\mathrm{fast}}\) and \(s_{p,\mathrm{slow}}\), respectively,
as selected by their mean runtimes on \(\mathcal{T}_{\mathrm{CC}}(p)\).
Comparing their exposure rates therefore shows whether generated efficiency tests 
simply follow this initial ranking or reveal additional bottlenecks.
Under Base, the \textit{Fast} group has a lower mean ASR than the
\textit{Slow} group, 41.32\% versus 59.92\%.
STAB raises both groups, but the relative increase is larger for \textit{Fast} evaluation solutions,
with average exposure rates of 59.35\% for \textit{Fast} and 74.01\% for \textit{Slow}.
As a result, the \textit{Fast}-\textit{Slow} gap is reduced by 21.18\% relative to Base. 
This reduced gap suggests that the larger separation under Base reflects limited bottleneck exposure rather 
than a robust efficiency difference between \textit{Fast} and \textit{Slow} evaluation solutions.

The \textit{Random} group provides a second check of this interpretation.
Because \(s_{p,\mathrm{slow}}\) is selected by mean runtime on \(\mathcal{T}_{\mathrm{CC}}(p)\),
a generated test set that merely preserves the benchmark ordering should 
keep \textit{Slow} above \textit{Random}.
Base does not consistently preserve this ordering: \textit{Random} exceeds \textit{Slow} 
in 7 of the 12 model and language pairs, even though 
its aggregate \textit{Random} average remains lower than \textit{Slow}.
Under STAB, the reversal becomes systematic: \textit{Random} exceeds \textit{Slow} in all 
12 model and language pairs, with average exposure rates of 76.62\% and 74.01\%.
Relative to Base, STAB improves the exposure rate more strongly for \textit{Random} 
than for \textit{Slow}, with relative increases of 36.48\% and 23.52\%, respectively.
This pattern indicates that the randomly sampled evaluation solutions contain latent runtime bottlenecks
that Base exposes only inconsistently, while STAB exposes them across all 12 pairs.
Together with the reduced gap between \textit{Fast} and \textit{Slow},
the reversal in which \textit{Random} exceeds \textit{Slow} shows that
STAB exposes bottlenecks more broadly and consistently across the five
evaluation solutions than Base. 

\begin{table*}[t]
\centering
\small
\caption{
ASR across five evaluation solutions \(s_{p,r}\).
Random$^\dagger$ reports the mean ASR over \(s_{p,r_i}\) for
\(i \in \{1,2,3\}\).
Avg. reports ASR averaged over five evaluation solutions and three languages.
Full results appear in Appendix~\ref{app:5_solution_analysis}.
}
\resizebox{\textwidth}{!}{%
\begin{tabular}{llcccccccccc}
\toprule
\multirow{2}{*}[-0.6ex]{Model} & \multirow{2}{*}[-0.6ex]{Method}
& \multicolumn{3}{c}{Python}
& \multicolumn{3}{c}{Java}
& \multicolumn{3}{c}{C++}
& \multirow{2}{*}[-0.6ex]{Avg.} \\
\cmidrule(lr){3-5}\cmidrule(lr){6-8}\cmidrule(lr){9-11}
& & Fast & Slow & Random$^\dagger$
  & Fast & Slow & Random$^\dagger$
  & Fast & Slow & Random$^\dagger$ & \\
\midrule
\multirow{2}{*}{Qwen-3.5}
& Base & 26.57\% & 57.10\% & 61.35\% & 27.60\% & 56.65\% & 59.43\% & 34.74\% & 57.19\% & 62.69\% & 54.02\% \\
& STAB & \textbf{56.14\%} & \textbf{74.20\%} & \textbf{74.56\%} & \textbf{59.46\%} & \textbf{75.11\%} & \textbf{77.41\%} & \textbf{59.33\%} & \textbf{69.11\%} & \textbf{72.27\%} & \textbf{71.07\%} \\
\midrule
\multirow{2}{*}{Gemma-4}
& Base & 42.32\% & 65.80\% & 41.77\% & 41.54\% & 57.38\% & 41.33\% & 58.07\% & 64.52\% & 41.16\% & 46.83\% \\
& STAB & \textbf{60.00\%} & \textbf{79.23\%} & \textbf{79.84\%} & \textbf{63.08\%} & \textbf{82.17\%} & \textbf{82.98\%} & \textbf{62.22\%} & \textbf{71.78\%} & \textbf{76.76\%} & \textbf{75.82\%} \\
\midrule
\multirow{2}{*}{Gemini-3.1}
& Base & 34.11\% & 54.88\% & 54.94\% & 35.48\% & 58.91\% & 59.79\% & 32.22\% & 42.52\% & 43.33\% & 48.82\% \\
& STAB & \textbf{59.23\%} & \textbf{75.65\%} & \textbf{77.84\%} & \textbf{59.82\%} & \textbf{77.65\%} & \textbf{79.91\%} & \textbf{58.52\%} & \textbf{67.11\%} & \textbf{73.93\%} & \textbf{72.87\%} \\
\midrule
\multirow{2}{*}{GPT-5.4}
& Base & 51.88\% & 68.21\% & 68.05\% & 57.47\% & 74.30\% & 73.82\% & 53.85\% & 61.63\% & 66.03\% & 66.07\% \\
& STAB & \textbf{56.04\%} & \textbf{73.91\%} & \textbf{74.81\%} & \textbf{58.01\%} & \textbf{75.66\%} & \textbf{77.31\%} & \textbf{60.30\%} & \textbf{66.59\%} & \textbf{71.83\%} & \textbf{70.83\%} \\
\bottomrule
\end{tabular}%
}
\label{tab:full_result}
\end{table*}

\subsection{Consistency Across Evaluation Solutions} \label{ssec:reference-variance}
A single algorithmic problem admits multiple accepted solutions in
\(\mathcal{A}(p)\) with different runtime profiles because the solutions can
use different algorithms, data structures, or implementation strategies.
The same efficiency test case can be adversarial for one evaluation solution
\(s_{p,r}\) but less adversarial for another.
A generator targeting an implementation artifact would perform strongly on
one evaluation solution and weaken when the evaluation aggregates over all
five solutions.
We use this variation as a consistency check by evaluating every
\(t \in \mathcal{T}_{\mathrm{gen}}(p)\) on the five evaluation solutions
\(s_{p,r}\) for each problem and implementation language.

Table~\ref{tab:full_result} reveals two consistent patterns across the five evaluation solutions.
First, STAB raises the bottleneck exposure rate over Base for the Fast, Slow, and Random groups.
The largest relative gain occurs for \(s_{p,\mathrm{fast}}\), whose mean ASR increases from 41.32\% to 59.35\%,
which is informative because fast solutions are the hardest to expose.
The mean ASR for \(s_{p,\mathrm{slow}}\) increases from 59.92\% to 74.01\%, and
the mean over the three randomly sampled evaluation solutions increases from 56.14\% to 76.62\%.
The improvement therefore does not concentrate on any evaluation-solution group.
Second, variation across the five evaluation solutions decreases on average.
The standard deviation over the three randomly sampled evaluation solutions \(s_{p,r_i}\) drops from 2.55\% 
under Base to 0.93\% under STAB, and the average range across the five evaluation solutions is reduced by 26.02\% relative to Base. 
The lower spread shows that STAB's gains do not favor any evaluation solution.

\subsection{Per-Module Contribution}\label{ssec:module-ablation}
The ablation compares STAB with three variants that remove the constraint saturator~(M1), the adversarial scenario injector~(M2), or both modules.
Removing M1 leaves the generator without a resolved valid boundary, whereas removing M2 leaves it without adversarial structural guidance.
Table~\ref{tab:ablation_result} shows that STAB achieves the highest ASR in all combinations.
Appendix~\ref{app:module-accuracy} reports the module-level accuracy of M1 and M2.


The single-module variants reveal complementary failure modes.
The variant without M1 can still use the construction principle matched by M2 but must infer the input size from the problem specification, 
and the inferred values fall below the largest admissible bound when constraints involve coupled variables or aggregate bounds.
The construction is therefore instantiated at a conservative scale that does not push the algorithm to its worst case.
The variant without M2 can use the resolved boundary from M1 but lacks the adversarial construction principle needed to convert 
that boundary into an effective efficiency test case.
The resulting input uses the boundary resolved by M1 but fills that boundary with a generic structure that does not expose the algorithmic bottleneck.
The variant without M2 outperforms the variant without M1 on both Gemini-3.1 and GPT-5.4 across Python, Java, and C++, 
suggesting that boundary resolution contributes more than adversarial construction when only one module is present.
STAB's additional gain over the variant without M2 nevertheless shows that adversarial construction principles are essential for fully effective efficiency test cases.
Only the full pipeline satisfies both conditions by instantiating the matched construction principle at the size resolved by M1.
Appendix~\ref{app:size_struct_analysis} compares the contributions of M1
and M2 across disjoint size-only and structure-only buckets and examines
how the two modules complement each other across bottleneck types.
Appendix~\ref{app:ablation_case_study} compares generators from STAB
and its two single-module variants on a shared problem, highlighting
differences in boundary use and adversarial structure.

\begin{table}[t]
\centering
\small
\setlength{\tabcolsep}{6pt}
\caption{
Ablation study on STAB.
M1 and M2 denote the constraint saturator and the adversarial scenario injector.
Each ASR value is averaged over the five evaluation solutions $s_{p,r}$.
}
\begin{tabular}{lccc}
\toprule
Method
 & $\mathrm{ASR}_{\mathrm{Python}}$
 & $\mathrm{ASR}_{\mathrm{Java}}$
 & $\mathrm{ASR}_{\mathrm{C++}}$ \\
\midrule
\multicolumn{4}{c}{\textit{Gemini-3.1}} \\
\midrule
STAB              & \textbf{73.68\%} & \textbf{75.44\%} & \textbf{69.48\%} \\
\quad $-$ M1      & 70.32\% & 73.67\% & 66.47\% \\
\quad $-$ M2      & 72.37\% & 75.40\% & 67.17\% \\
\quad $-$ M1, M2  & 50.76\% & 54.75\% & 40.95\% \\
\midrule
\multicolumn{4}{c}{\textit{GPT-5.4}} \\
\midrule
STAB              & \textbf{70.88\%} & \textbf{73.12\%} & \textbf{68.48\%} \\
\quad $-$ M1      & 64.15\% & 68.45\% & 61.51\% \\
\quad $-$ M2      & 66.45\% & 70.19\% & 64.07\% \\
\quad $-$ M1, M2  & 64.85\% & 70.64\% & 62.71\% \\
\bottomrule
\end{tabular}
\label{tab:ablation_result}
\end{table}

\subsection{Comparison with Prior Methods}\label{ssec:prior-comparison}

Table~\ref{tab:prior_comparison} compares STAB with EvalPerf and WEDGE,
both of which require a reference implementation.
STAB achieves the highest ASR in all 12 model-language combinations,
showing that its advantage extends to both open-source and closed-source models.
Averaged across four LLMs and three implementation languages, STAB achieves
72.65\% ASR, compared with 35.37\% for EvalPerf and 37.35\% for WEDGE.
The relative gains are 105.39\% over EvalPerf and 94.51\% over WEDGE.

The results indicate that the gap is not explained by access to a reference implementation, 
but by the kind of signal each method uses to construct test cases.
EvalPerf remains effective when the worst case is mainly scale-driven,
but its average ASR is 35.37\% because increasing a size parameter does not specify
the ordering, topology, or coupled dimensions needed by structural worst cases.
WEDGE also uses a reference implementation, but its search signal is tied
to the execution behavior of that implementation.
Such a signal can find slow paths in the reference implementation, but it is
less reliable when the bottleneck depends on the shape of the whole input,
such as a reverse-sorted array, a bamboo tree, or a dense graph that maximizes
both vertices and edges under the input constraints.
STAB avoids these two failure modes by combining a resolved boundary derived
from the problem specification with retrieved construction principles.
This combination explains why STAB outperforms both methods requiring a
reference implementation despite using only the problem specification at
inference.
Appendix~\ref{app:prior_work_case_study} further analyzes these failure modes
in the reference implementation setting.
Appendix~\ref{app:code-based-baselines} additionally compares STAB with
HARDTESTGEN and CodeContests-O, two code-based methods for
correctness-oriented test generation.
STAB's advantage extends to both efficiency-oriented and correctness-oriented test-generation methods.

\begin{table}[t]
\centering
\small
\setlength{\tabcolsep}{6pt}
\caption{
Comparison with prior efficiency-oriented generators across four models.
Each ASR value is averaged over the five evaluation solutions $s_{p,r}$
per problem and implementation language.
}
\begin{tabular}{llccc}
\toprule
Model & Method & $\mathrm{ASR}_{\mathrm{Python}}$ & $\mathrm{ASR}_{\mathrm{Java}}$
& $\mathrm{ASR}_{\mathrm{C++}}$ \\
\midrule
\multirow{3}{*}{Qwen-3.5}
 & EvalPerf & 29.72\% & 34.64\% & 31.72\% \\
 & WEDGE    & 28.29\% & 33.77\% & 40.13\% \\
 & STAB     & \textbf{70.80\%} & \textbf{73.36\%} & \textbf{69.05\%} \\
\midrule
\multirow{3}{*}{Gemma-4}
 & EvalPerf & 36.41\% & 39.27\% & 33.41\% \\
 & WEDGE    & 31.39\% & 40.83\% & 54.98\% \\
 & STAB     & \textbf{75.75\%} & \textbf{78.84\%} & \textbf{72.86\%} \\
\midrule
\multirow{3}{*}{Gemini-3.1}
 & EvalPerf & 28.91\% & 31.89\% & 23.21\% \\
 & WEDGE    & 35.79\% & 41.09\% & 38.92\% \\
 & STAB     & \textbf{73.68\%} & \textbf{75.44\%} & \textbf{69.48\%} \\
\midrule
\multirow{3}{*}{GPT-5.4}
 & EvalPerf & 45.35\% & 47.84\% & 42.06\% \\
 & WEDGE    & 29.00\% & 36.18\% & 37.81\% \\
 & STAB     & \textbf{70.88\%} & \textbf{73.12\%} & \textbf{68.48\%} \\
\bottomrule
\end{tabular}
\label{tab:prior_comparison}
\end{table}

\subsection{Runtime Discrimination Analysis}\label{ssec:worst-case-comparison}

\begin{figure*}[h]
    \centering
    \includegraphics[width=\textwidth]{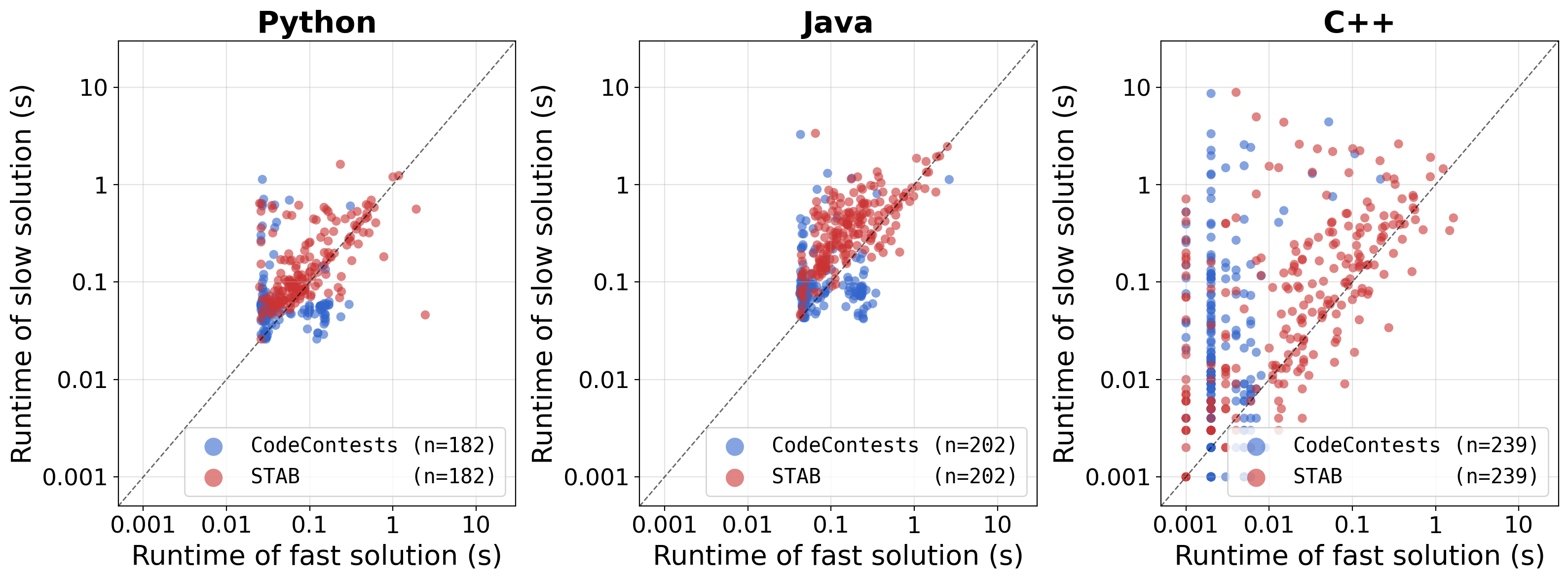}
    
    \caption{
    Runtime comparison of test cases selected from
    \(\mathcal{T}_{\mathrm{CC}}(p)\)~(blue) and
    \(\mathcal{T}_{\mathrm{gen}}(p)\)~(red) on
    \(s_{p,\mathrm{fast}}\) and \(s_{p,\mathrm{slow}}\) for Gemma-4.
    Points above the diagonal indicate that
    \(s_{p,\mathrm{slow}}\) runs longer than \(s_{p,\mathrm{fast}}\).
    }
    \label{fig:fast_slow_runtime}
\end{figure*}

We use Gemma-4 because it achieves the highest average ASR across Python,
Java, and C++ in Table~\ref{tab:full_result}.
For each problem \(p\), we independently select one test case from
\(\mathcal{T}_{\mathrm{CC}}(p)\) and one from
\(\mathcal{T}_{\mathrm{gen}}(p)\).
Within each test set, the selected test case maximizes
\(\max\left\{\tau(s_{p,\mathrm{fast}},t),\tau(s_{p,\mathrm{slow}},t)\right\}\).
We then measure the runtime of the selected test case on both
\(s_{p,\mathrm{fast}}\) and \(s_{p,\mathrm{slow}}\).
Figure~\ref{fig:fast_slow_runtime} plots the resulting runtime pairs on
log-scaled axes.
Points above the diagonal indicate test cases for which
\(s_{p,\mathrm{slow}}\) runs longer than \(s_{p,\mathrm{fast}}\).
Across Python, Java, and C++, STAB test cases widen the gap between fast and slow solutions.
On Java, STAB increases the median fast runtime by 198\% and the median slow runtime 
by 322\% relative to the CodeContests test cases, 
and the share of problems where the slow solution takes longer than 
the fast solution rises from 69.8\% to 85.1\%.
On C++, the CodeContests test cases leave 54.4\% of problems in a very low-runtime regime, 
where the fast and slow solutions are difficult to distinguish by execution time.
STAB reduces this fraction to 15.9\%, making the runtime gap between 
fast and slow solutions more visible.
The widened gap does not by itself imply that any specific solution is 
algorithmically superior.
It indicates that STAB test cases make algorithmic complexity differences 
observable in measured runtime, supporting downstream tasks 
such as solution ranking.

\section{Conclusion}\label{sec:conclusion}
We present \textbf{STAB}, a specification-driven pipeline that generates efficiency test 
inputs from the problem specification alone.
STAB combines constraint saturation for admissible boundary assignments with 
adversarial scenario retrieval from a 13-scenario catalog.
On CodeContests, STAB outperforms standard prompting across three languages 
and four LLMs, and surpasses prior efficiency-oriented generators in the evaluated 
settings, all without using a reference implementation.
Appendix~\ref{app:limitation} discusses limitations of the scenario
catalog and constraint parser.

\bibliographystyle{iclr2027_conference}
\bibliography{custom}

\clearpage

\appendix

\section{Implementation Details} \label{app:impl_detail}

We evaluate 270 CodeContests problems after filtering, consisting of 157 \texttt{test} problems and 113 \texttt{valid} problems.
For scenario retrieval, we build the KNN anchor pool from 5,306 training problems matched by the 13-scenario catalog's keyword matcher.
All experiments were conducted using three NVIDIA RTX PRO 6000 GPUs.
For LLM sampling, we used temperature \(=1.0\), top-\(p=0.95\), and a maximum output length of 4096 tokens.
For each problem, we generated five efficiency test cases.
Generated test cases were judged in a containerized DOMjudge environment to measure runtime for each solution and language.
All solutions were compiled and executed inside DOMjudge judgehosts under the time limit specified for each CodeContests problem.

\section{Detailed Dataset Construction}\label{app:dataset}

We use the CodeContests dataset~\citep{abs-2203-07814}.
The CodeContests evaluation splits contain 165 \texttt{test} problems and 117 \texttt{valid} problems.
We first executed candidate accepted solutions on the CodeContests test cases 
and excluded problems for which at least one evaluation solution failed.
This filtering removed 8 \texttt{test} problems and 4 \texttt{valid} problems, leaving 
157 \texttt{test} problems and 113 \texttt{valid} problems, for a total of 270 evaluated 
problems. This filtering ensures that runtime thresholds are computed only for problems 
whose evaluation solutions pass \(\mathcal{T}_{\mathrm{CC}}(p)\).
The CodeContests training split contains 13,328 problems.
We do not use the training split for runtime evaluation.
Instead, we use it only to build the KNN anchor pool for the adversarial scenario injector.
We apply the 13-scenario catalog's keyword matcher to the training problem descriptions 
and retain 5,306 matched problems, covering 39.8\% of the training split.
This anchor construction uses only problem specifications and does not use
solution code, generated test cases, runtime measurements, or evaluation
outcomes.


Solutions in CodeContests include multiple implementation languages, including
Python~2.
We excluded Python~2 because it is an End-of-Life environment without official
support.
Accordingly, we evaluated Python~3, Java, and C++ under the problem-specific
time limits provided by CodeContests.
For each problem \(p\) and implementation language, we measured the mean runtime
of every \(s \in \mathcal{A}(p)\) on
\(\mathcal{T}_{\mathrm{CC}}(p)\).
We then selected \(s_{p,\mathrm{fast}}\), \(s_{p,\mathrm{slow}}\), and
\(s_{p,r_i}\) for \(i \in \{1,2,3\}\) as the five evaluation solutions.
All five evaluation solutions passed \(\mathcal{T}_{\mathrm{CC}}(p)\) under
the DOMjudge environment.
We ran \(\mathcal{T}_{\mathrm{gen}}(p)\) on these evaluation solutions to
identify runtime bottlenecks not exposed by
\(\mathcal{T}_{\mathrm{CC}}(p)\).

\section{DOMjudge Execution Environment}\label{app:domjudge}
General operating system environments or simple sandboxes can introduce runtime variance 
across executions due to background processes, CPU scheduling fluctuations, 
and hardware cache effects.
DOMjudge judgehosts execute submissions under controlled restrictions, using cgroups for 
process isolation, memory accounting, network restrictions, and CPU binding.
DOMjudge enforces time limits primarily using CPU time, with a hard wall-clock limit as 
a secondary guard, and reports runtime and memory usage under this judgehost environment.

Consequently, this controlled environment reduces distortions from system noise while 
preserving language-specific runtime behavior.
It provides a stable basis for measuring the time costs imposed by
\(\mathcal{T}_{\mathrm{gen}}(p)\) on the evaluation solutions \(s_{p,r}\).


\section{Comparison with Additional Code-Based Methods}\label{app:code-based-baselines}
We evaluate HARDTESTGEN~\citep{he2026hardtestgen} and
CodeContests-O~\citep{cai2026codecontests}, which use problem-specific code
primarily to distinguish correct programs from rejectable ones.
This comparison examines whether correctness-oriented objectives transfer
to runtime bottleneck exposure.
Table~\ref{tab:code-based-baselines} shows that STAB achieves the highest
ASR in all three implementation languages.
Averaged across Python, Java, and C++, HARDTESTGEN, CodeContests-O, and STAB
achieve 22.85\%, 47.78\%, and 70.83\% ASR, respectively.
The gap is not primarily attributable to invalid inputs.
HARDTESTGEN and CodeContests-O generate test cases that satisfy the problem
constraints in 99.7\% and 93.0\% of cases, respectively, and the resulting
inputs execute correctly on the evaluation solutions \(s_{p,r}\).
Their lower ASR instead reflects the objective used to select test cases.

HARDTESTGEN targets failures in brute-force or otherwise rejectable programs,
including failures caused by exceeding the time limit.
A test case is successful when it causes such a program to time out or
distinguishes it from a correct program.
Its runtime signal therefore separates rejectable programs from correct
programs rather than maximizing the slowdown of an evaluation solution that
already satisfies the time limit.

CodeContests-O optimizes correctness discrimination through iterative feedback
and does not include an explicit runtime objective.
Its objective can favor larger inputs because an inefficient incorrect program
can behave like a correct program on small test cases and diverge only when the
input becomes sufficiently large.
This indirect pressure helps explain why CodeContests-O achieves higher ASR
than HARDTESTGEN.
Once a test case distinguishes the correct and incorrect programs, however,
the objective provides no incentive to increase the runtime of the correct
program or to construct the problem-level adversarial structure needed to
expose its runtime bottleneck.

ASR evaluates a different target.
It runs each generated efficiency test case on evaluation solutions that
already satisfy the time limit and checks whether the test exceeds the
CodeContests runtime threshold for the same evaluation solution \(s_{p,r}\).
A valid test case can therefore provide sufficient correctness discrimination
while imposing limited runtime pressure on an evaluation solution.
The results show that the correctness-oriented objectives of HARDTESTGEN and
CodeContests-O transfer only partially to runtime bottleneck exposure.
STAB achieves higher ASR while remaining the only method in this comparison
that does not consume a problem-specific implementation.

\begin{table}[h]
\centering
\small
\setlength{\tabcolsep}{4pt}
\caption{Comparison with additional code-based stress-test generation
methods on GPT-5.4.}
\begin{tabular}{lccc}
\toprule
Method & $\mathrm{ASR}_{\mathrm{Python}}$ & $\mathrm{ASR}_{\mathrm{Java}}$ & $\mathrm{ASR}_{\mathrm{C++}}$ \\
\midrule
HARDTESTGEN    & 6.18\%  & 37.18\% & 25.19\% \\
CodeContests-O & 43.25\% & 52.13\% & 47.97\% \\
\midrule
\textbf{STAB}  & \textbf{70.88\%} & \textbf{73.12\%} & \textbf{68.48\%} \\
\bottomrule
\end{tabular}
\label{tab:code-based-baselines}
\end{table}

\section{Retrieval Strategy Ablation}\label{app:retrieval-ablation}
We evaluated eight retrieval configurations on GPT-5.4 using 50 randomly
sampled problems. For each problem and implementation language, generated
tests were evaluated on \(s_{p,\mathrm{fast}}\) and
\(s_{p,\mathrm{slow}}\). All configurations shared the same problem
specification and M1-resolved boundary; only the retrieval output supplied by
M2 differed. \textit{No retrieval} supplied no construction principle,
\textit{Full catalog} supplied the complete scenario catalog, and
\textit{Keyword only} supplied no scenario when no keyword matched.
\textit{Union} injected the union of Keyword Match and KNN outputs, whereas
STAB uses Keyword Match when available and invokes KNN only otherwise.

Table~\ref{tab:retrieval-ablation} shows that selective retrieval is more
effective than unrestricted catalog exposure. Averaged across the three
implementation languages, STAB achieves 76.24\% ASR, compared with 73.71\%
for Union, 71.34\% for the strongest KNN configuration (\(k=2\)), 68.98\%
for No retrieval, and 67.07\% for Full catalog. Keyword only records 52.53\%,
which is consistent with specifications that imply an algorithmic structure
without naming it explicitly. Among the KNN configurations, \(k=2\) performs
best. The current design uses a fixed \(k=2\) and no similarity threshold.

\begin{table}[h]
\centering
\small
\setlength{\tabcolsep}{6pt}
\caption{
Ablation of eight M2 retrieval configurations on GPT-5.4 using 50
randomly sampled problems.
}
\begin{tabular}{lccc}
\toprule
Retrieval & $\mathrm{ASR}_{\mathrm{Python}}$ & $\mathrm{ASR}_{\mathrm{Java}}$
& $\mathrm{ASR}_{\mathrm{C++}}$ \\
\midrule
No retrieval & 65.71\% & 72.43\% & 68.80\% \\
Full catalog & 58.00\% & 71.62\% & 71.60\% \\
Keyword only & 46.57\% & 56.22\% & 54.80\% \\
KNN ($k{=}1$) & 64.57\% & 72.70\% & 72.00\% \\
KNN ($k{=}2$) & 65.14\% & 74.86\% & 74.00\% \\
KNN ($k{=}5$) & 62.57\% & 71.08\% & 73.00\% \\
Union & 69.71\% & 77.03\% & 74.40\% \\
\midrule
\textbf{STAB} & \textbf{72.86\%} & \textbf{80.27\%} & \textbf{75.60\%} \\
\bottomrule
\end{tabular}
\label{tab:retrieval-ablation}
\end{table}

\section{Size-Only vs.\ Structure-Only Buckets} \label{app:size_struct_analysis}
We conduct a bucketed ablation to examine whether M1's larger
contribution reflects a benchmark dominated by size-driven bottlenecks.
A problem is assigned to the size-only or structure-only bucket when its
matched implementations belong exclusively to the Size Only or Structural
class, respectively.
The former captures problems for which input volume is sufficient to increase
runtime, whereas the latter requires an order, shape, or topology in
addition to scale.
Excluding numerical and mixed-class problems leaves 25 size-only and
33 structure-only problems.
Table~\ref{tab:bucket-ablation} compares STAB with the variants that remove M1
or M2 on GPT-5.4 and Gemini-3.1, averaging ASR over three implementation
languages and five evaluation solutions.

M1's contribution is not concentrated in size-only problems.
Removing M1 produces a larger relative ASR decrease in the structure-only
bucket than in the size-only bucket: 18.46\% versus 13.93\% for GPT-5.4
and 4.84\% versus 1.97\% for Gemini-3.1.
The Structural label indicates that scale alone is insufficient, not that scale is irrelevant.
A reverse-sorted sequence or a bamboo tree still requires enough elements or
vertices for its structural worst case to produce an observable runtime effect.
M1 therefore supplies the resolved boundary needed to instantiate the
construction principle from M2 at sufficient scale.

M2 also contributes more strongly to structure-only problems.
Removing M2 decreases ASR by 11.00\% and 5.69\% in the GPT-5.4
structure-only and size-only buckets, respectively; the corresponding decreases
for Gemini-3.1 are 5.35\% and 0.59\%.
In size-only problems, reaching the resolved boundary increases the workload
regardless of input arrangement.
In structure-only problems, a maximum-size input can still miss the slow path
when its order or topology is generic.
STAB achieves the highest ASR in all four model--bucket combinations,
confirming that M1 determines the admissible scale while M2 determines how
that scale is structured to expose the bottleneck.

\begin{table}[h]
\centering
\small
\setlength{\tabcolsep}{4pt}
\caption{
Size-only vs.\ structure-only ablation.
}
\begin{tabular}{llccc}
\toprule
Model & Bucket & $-$M1 & $-$M2 & STAB \\
\midrule
\multirow{2}{*}{GPT-5.4}
 & size-only      & 65.47\% & 71.74\% & \textbf{76.07\%} \\
 & structure-only & 63.58\% & 69.39\% & \textbf{77.97\%} \\
\cmidrule(lr){1-5}
\multirow{2}{*}{Gemini-3.1}
 & size-only      & 77.05\% & 78.14\% & \textbf{78.60\%} \\
 & structure-only & 77.89\% & 77.47\% & \textbf{81.85\%} \\
\bottomrule
\end{tabular}
\label{tab:bucket-ablation}
\end{table}

\begin{table}[t]
\centering
\small
\caption{C++ ASR of prior efficiency-oriented generators under the reference implementation setting.}
\begin{tabular}{llc}
\toprule
Model & Method & $\mathrm{ASR}_{\mathrm{C++}}$ \\
\midrule
\multirow{2}{*}{Gemini-3.1}
 & EvalPerf & 22.52\% \\
 & WEDGE    & 43.21\% \\
\cmidrule(lr){1-3}
\multirow{2}{*}{GPT-5.4}
 & EvalPerf & 35.65\% \\
 & WEDGE    & 41.83\% \\
\bottomrule
\end{tabular}
\label{tab:prior_input_solution}
\end{table}

\section{Reference-Implementation Evaluation}\label{app:prior_work_case_study}
Table~\ref{tab:prior_input_solution} examines prior methods in their most favorable setting:
a C++ reference implementation is provided as input, and the generated test cases are evaluated 
on that same reference implementation.
This setting allows EvalPerf and WEDGE to specialize their test cases to the code used during evaluation.
However, their C++ ASR remains limited even under this reference implementation setting.
On Gemini-3.1, EvalPerf reaches 22.52\% ASR and WEDGE reaches 43.21\%.
On GPT-5.4, the corresponding values are 35.65\% and 41.83\%.
These results indicate that access to the evaluated reference implementation does not by itself produce 
test cases that consistently expose efficiency bottlenecks.

The reference implementation setting reveals two failure modes.
First, EvalPerf is mainly driven by input scaling.
Because this scaling is not paired with explicit constraint resolution or an adversarial construction principle, 
the generated test cases can violate coupled constraints or remain large but structurally ordinary.
Second, WEDGE derives slowdown conditions from executions of a reference implementation.
This signal is strongest when the benchmark executions reveal a clear fast-slow runtime contrast, 
but it becomes weak when such a contrast is absent.
Thus, although WEDGE's conditions can in principle be more targeted than EvalPerf's, their reliance on a specific reference implementation prevents 
consistent recovery of the adversarial construction principle needed to expose the algorithmic bottleneck.
STAB avoids these failures by fixing both parts before generator synthesis: a validated boundary from the constraint saturator and an adversarial construction principle from the adversarial scenario injector.
The LLM therefore synthesizes a generator from an already constrained and structured specification, rather than relying on input scaling or execution-derived slowdown signals.
Together, these results support the main comparison: prior methods fall behind because they miss the combination of valid boundary resolution and adversarial construction principles that STAB encodes explicitly.

\section{Case Study: Complementary Modules}\label{app:ablation_case_study}

We present a generator-level case study on 1582\_F1, \textit{Korney Korneevich and XOR}, 
to show how the two STAB modules provide complementary information.
The problem gives an array \(a\) of length \(n\), where \(1 \le n \le 10^5\) and \(0 \le a_i \le 500\), 
and asks for every value \(x \ge 0\) that can be obtained as the bitwise XOR of a strictly increasing subsequence of \(a\).
Here, increasing refers to the selected values rather than their positions.
A valid subsequence must therefore satisfy
\(i_1 < i_2 < \cdots < i_m\) and
\(a_{i_1} < a_{i_2} < \cdots < a_{i_m}\).
Common solutions track which XOR values are reachable together with the smallest possible last selected value.
Therefore, a strong efficiency test must satisfy two conditions:
the input length \(n\) must be large enough to increase the number of processed elements, and the value distribution must activate a large portion of the \((\text{last\_value}, \text{current\_xor})\) state space.
This makes the problem a balanced case in which boundary resolution and adversarial scenario guidance are both necessary.

Figure~\ref{fig:case_study_stab} shows how the STAB generator combines the
boundary resolved by the constraint saturator with the construction principle
supplied by the adversarial scenario injector.
The constraint saturator resolves \(n=100{,}000\), saturating the valid boundary.
Within this boundary, the generator first repeats a 502-element cycle formed
by the increasing triples \((t,2t,3t)\) for \(1 \le t \le 166\), followed by
the extreme-value pattern \((0,500,1,499)\).
This base cycle contains 335 distinct values.
The generator then replaces the first 2,000 positions of every 10,000-element
block with a consecutive sequence modulo 501, shifting its starting value by
83 between blocks.
Each replaced segment therefore traverses the full admissible value range.
The resulting array combines repeated increasing transitions, alternating
extreme values, and shifted full-range sweeps at the maximum valid length.
This structure provides repeated opportunities to extend strictly increasing
subsequences and update XOR states across the value range.
In the case-study evaluation, the generated test exceeded the CodeContests
runtime threshold for all five evaluation solutions, including the solution
not exposed by the generator using only the constraint saturator.

Figure~\ref{fig:case_study_m1_only} shows the variant without the adversarial
scenario injector.
This variant keeps the resolved boundary but loses the adversarial value
structure.
It also uses \(n=100{,}000\), but divides the array into 20 constant blocks drawn
from only 11 distinct values.
These long constant regions provide substantially fewer increasing transitions
than the repeated triples and full-range sweeps used by STAB.
The resulting value distribution activates the reachable XOR states less
aggressively, even though both generators use the same input length.
Consequently, this test does not drive all five evaluation solutions beyond
their CodeContests runtime thresholds.
This result shows that boundary saturation alone is insufficient when the
values within that boundary have limited diversity and structure.

\begin{figure}[h]
\centering
\begin{stabcode}{STAB generator for 1582\_F1}
def generate() -> str:
    n = 100000
    cycle = []
    for t in range(1, 167):
        cycle.extend([t % 501, (2*t) % 501, (3*t) % 501])
    cycle.extend([0, 500, 1, 499])
    a = []
    while len(a) < n: a.extend(cycle)
    a = a[:n]
    inc = 0
    for k in range(0, n, 10000):
        for j in range(2000):
            if k + j < n: a[k+j] = (inc + j) % 501
        inc = (inc + 83) % 501
    return f"{n}\n" + " ".join(map(str, a))
\end{stabcode}
\caption{
Generator synthesized by GPT-5.4 using STAB for 1582\_F1, \textit{Korney Korneevich and XOR}.
}
\label{fig:case_study_stab}
\end{figure}

\begin{figure}[h]
\centering
\begin{stabcode}{Constraint Saturator generator for 1582\_F1}
def generate() -> str:
    n = 100000
    values = [0, 1, 2, 3, 4, 5, 6, 7, 8, 9, 10]
    blocks = 20
    per_block = n // blocks
    a = []
    for b in range(blocks):
        a.extend([values[b % len(values)]] * per_block)
    a.extend([values[-1]] * (n - len(a)))
    return "\n".join([str(n), " ".join(map(str, a))])
\end{stabcode}
\caption{
Generator synthesized by GPT-5.4 using only the constraint saturator for 1582\_F1.
}
\label{fig:case_study_m1_only}
\end{figure}

\begin{figure}[h]
\centering
\begin{stabcode}{Adversarial Scenario Injector generator for 1582\_F1}
def generate() -> str:
    n = 20
    a = [(i * 7) % 31 for i in range(n)]
    return "\n".join([str(n), " ".join(map(str, a))])
\end{stabcode}
\caption{
Generator synthesized by GPT-5.4 using only the adversarial scenario injector for 1582\_F1.
}
\label{fig:case_study_m2_only}
\end{figure}

Figure~\ref{fig:case_study_m2_only} shows the variant without the constraint
saturator.
This variant exhibits the opposite failure mode.
Without the resolved boundary, it infers \(n=20\) from the problem specification.
Given this inferred scale, the generator uses an XOR-diverse construction, 
\((i \cdot 7) \bmod 31\), guided by the adversarial scenario injector.
This construction uses the small inferred length effectively: since 7 is coprime to 31, 
the first 20 values are distinct and avoid trivial repetition.
However, the inferred length is \(5{,}000\) times smaller than the valid upper bound \(10^5\).
The construction is reasonable within the inferred scale, but the input is too small to create enough outer loop work.
This result shows that adversarial construction alone remains under-scaled without boundary resolution.

This case study highlights the complementary roles of the two components.
Without the adversarial scenario injector, the generator reaches the valid
boundary but fills it with low-diversity constant blocks.
Without the constraint saturator, the generator produces a diverse value
sequence but instantiates it at \(n=20\), which is \(5{,}000\) times smaller
than the valid upper bound.
STAB combines the maximum valid length with increasing triples, extreme-value
alternation, and shifted full-range sweeps.
The resulting test case is therefore both large and structurally adversarial.

\section{Module-Level Accuracy}\label{app:module-accuracy}
Table~\ref{tab:module-accuracy} directly evaluates whether M1 and M2
produce correct intermediate outputs.
For M1, we compared the extracted and resolved boundaries with annotated
boundary constraints at the constraint level.
M1 achieves 90.66\% accuracy.
The remaining errors consist of omitted prose-only constraints~(5.3\%),
unsupported notation patterns~(2.9\%), and parsing errors~(1.1\%).
M1 uses a regex-based parser rather than an LLM.
Constraints omitted from the resolved boundary remain available in the
problem specification supplied to the generator, while the conservative parser
avoids injecting unsupported bounds as boundary guidance.

M2 is evaluated at the problem level.
Keyword Match runs first, and KNN is used only when Keyword Match retrieves no scenario.
A retrieval is correct when at least one retrieved scenario belongs to the annotated relevant-scenario set.
M2 achieves 72.59\% overall accuracy, with 70.40\% for Keyword Match and 74.48\% for KNN.
Keyword Match errors mainly arise from lexical overlap between related
scenarios, whereas KNN errors occur more frequently for scenarios represented by fewer anchors.
These evaluations complement the end-to-end ASR ablation by directly
characterizing the reliability and failure modes of both modules.

\begin{table}[h]
\centering
\small
\caption{
Module-level accuracy against ground truth.
M1 is evaluated per constraint and M2 per problem.
}
\begin{tabular}{lc}
\toprule
Component & Accuracy \\
\midrule
Module 1 (Constraint Saturator) & 90.66\% \\
\midrule
Module 2 (Adversarial Scenario Injector) & 72.59\% \\
\quad (a) Keyword Match & 70.40\% \\
\quad (b) KNN & 74.48\% \\
\bottomrule
\end{tabular}
\label{tab:module-accuracy}
\end{table}

\section{Limitations}\label{app:limitation}
STAB is designed to make efficiency test case generation for algorithmic bottlenecks both
reproducible and specification-driven, but several practical constraints
define its current scope and point to natural extensions.

\paragraph{Coverage of the scenario catalog.}
The adversarial scenario catalog spans 13 scenarios that correspond to algorithm families
commonly encountered in competitive programming.
The adversarial scenario injector retrieves multiple matching scenarios per problem and 
encodes their construction principles into the structured generation specification,
but two cases remain outside its current coverage.
Algorithm families not represented in the catalog have no corresponding construction 
principle to retrieve.
Composite problems whose worst-case behavior emerges from interactions between scenarios 
receive only the union of individual construction principles,
rather than an interaction-specific principle that captures the combined bottleneck.
Both gaps require manual extension of the catalog or its interaction structure,
which the current pipeline does not automate.

\paragraph{Robustness of the constraint parser.}
The constraint saturator uses a regular expression parser designed for common constraint 
formats, including ranges, chained inequalities, product bounds, and aggregate size bounds.
This parser covers the constraint expressions that appear frequently in CodeContests, 
but it does not cover every way constraints can be written.
Constraints with deeply nested conditions, implicit dependencies across variables,
or prose-only restrictions can remain difficult to extract with regular expressions.
Extending STAB to benchmarks with more irregular constraint descriptions requires
a richer parser or an LLM-assisted constraint extraction step.

\section{Execution Validity of STAB-Generated Test Cases}\label{app:execution-validity}
We separately evaluated the execution validity of the efficiency test
cases generated by STAB.
For each execution record, we first ran the corresponding accepted
evaluation solution on the generated input to obtain the expected output.
We then evaluated the solution in DOMjudge using the resulting input--output pair.
We categorized each execution into three outcomes:
\emph{Validity} for Accepted~(AC), \emph{Timeout} for time limit
exceeded~(TLE), and \emph{Invalid} for wrong answer or runtime error~(WA/RE).
Timeout was reported separately because TLE reflects excessive runtime
rather than an incorrect output or runtime error.

Table~\ref{tab:execution-validity} reports the execution outcomes for
the four LLMs.
Each percentage is computed over the evaluated execution records for
the corresponding LLM.
Validity ranges from 95.42\% for Gemini-3.1 to 97.06\% for Gemma-4,
while Invalid accounts for 2.68--4.20\% of the evaluated executions.
Timeout accounts for 0.26--0.80\% of the executions.
Thus, 95.42--97.06\% of the evaluated executions complete successfully
and match the expected output, whereas 2.68--4.20\% result in WA or RE.
Execution validity is reported separately from ASR because the two
evaluations characterize different properties.
These results demonstrate that STAB achieves consistently
high execution validity across four LLMs.

\begin{table}[h]
\centering
\small
\setlength{\tabcolsep}{10pt}
\renewcommand{\arraystretch}{1.1}
\caption{Execution validity evaluation of STAB-generated test cases.}
\label{tab:execution-validity}
\begin{tabular}{lccc}
\toprule
Model & Validity~(AC) & Timeout~(TLE) & Invalid~(WA/RE) \\
\midrule
Qwen-3.5   & 95.43\% & 0.37\% & 4.20\% \\
Gemma-4    & 97.06\% & 0.26\% & 2.68\% \\
Gemini-3.1 & 95.42\% & 0.53\% & 4.05\% \\
GPT-5.4    & 95.75\% & 0.80\% & 3.45\% \\
\bottomrule
\end{tabular}
\end{table}

\section{Per-Solution Performance Analysis}\label{app:5_solution_analysis}
Table~\ref{tab:solution_variance} reports ASR separately for the five
evaluation solutions \(s_{p,r}\) used for each problem and implementation
language.
The Fast and Slow columns correspond to \(s_{p,\mathrm{fast}}\) and
\(s_{p,\mathrm{slow}}\), which have the lowest and highest mean runtimes on
\(\mathcal{T}_{\mathrm{CC}}(p)\), respectively.
The R1, R2, and R3 columns correspond to the three randomly sampled
evaluation solutions \(s_{p,r_i}\) for \(i \in \{1,2,3\}\).
The Fast and Slow labels therefore describe the two endpoints of the runtime
ranking induced by the CodeContests test set rather than fixed categories of
algorithmic quality.
Unlike the Random column in Table~\ref{tab:full_result}, which averages the
three randomly sampled solutions, Table~\ref{tab:solution_variance} preserves
their individual ASR values.
This disaggregation reveals whether the aggregate Random result is stable
across R1, R2, and R3 or is disproportionately influenced by one sampled
evaluation solution.
It also shows whether the relative performance of each method is preserved
across evaluation solutions with different runtime profiles.
Consistent results across all five columns provide stronger evidence that the
generated tests expose problem-level bottlenecks rather than behavior specific
to one evaluation solution.

Each ASR column uses the solution-specific threshold
\(\tau_{p,r}^{*}\), defined as the largest runtime induced by
\(\mathcal{T}_{\mathrm{CC}}(p)\) on the same evaluation solution.
A generated efficiency test case is counted as successful for
\(s_{p,r}\) only when its runtime exceeds this solution-specific benchmark threshold.
Fast, Slow, R1, R2, and R3 are therefore not evaluated against a shared absolute runtime cutoff.
This solution-specific thresholding accounts for the distinct runtime profile
of each evaluation solution and prevents differences in raw execution speed
from determining the success criterion.
Consequently, each reported value measures how frequently the generated
efficiency test cases induce runtimes above the maximum observed on the
CodeContests test set for the same solution, rather than comparing raw
execution times across heterogeneous implementations.

The Avg. column reports the arithmetic mean ASR across Fast, Slow, R1, R2,
and R3, assigning equal weight to the five evaluation solutions.
The Range column reports the difference between the maximum and minimum ASR
within the same model--language--method combination.
Avg. therefore measures the overall strength of bottleneck exposure, whereas
Range measures sensitivity to the choice of evaluation solution.
Range alone does not indicate effectiveness: a narrow Range can arise when
all five ASR values are uniformly low, whereas a wider Range can arise even
when all five values increase by different amounts.
Because Range remains unchanged when all five ASR values increase by the same
amount, it does not capture a uniform upward shift in performance.
Avg. captures this shift, whereas Range records only the spread between the
highest and lowest ASR values.
A higher Avg. therefore indicates stronger average bottleneck exposure even
when the corresponding Range is wider, provided that all five reported ASR
values also increase.

STAB achieves a higher Avg. than Base in all 12 model--language combinations
and a higher ASR in 59 of the 60 aggregated
model--language--evaluation-solution comparisons.
The sole exception is R1 for Qwen-3.5 on C++, where ASR decreases from
73.19\% under Base to 71.70\% under STAB, corresponding to a 2.04\%
relative decrease.
In the same model--language combination, however, Avg. increases from
56.00\% to 69.05\%, a 23.30\% relative increase, while Range decreases by
65.12\% relative to Base.
STAB has a wider Range in four combinations: GPT-5.4 on Python, Gemma-4 and
GPT-5.4 on Java, and Gemini-3.1 on C++.
Even in these combinations, Avg. increases by 9.30\%, 76.85\%, 3.51\%, and
69.67\%, respectively, relative to Base, and STAB exceeds Base for each of
Fast, Slow, R1, R2, and R3.
The wider Range in these four combinations therefore reflects differences in
the magnitude of improvement across evaluation solutions rather than a
decrease in any of the five reported ASR values.
All five values remain above their Base counterparts in these combinations,
so the broader spread is not caused by a loss on a particular evaluation solution.
Across all 12 combinations, STAB produces a smaller Range in eight and
reduces the average Range by 26.02\% relative to Base.
The mean within-cell standard deviation across R1, R2, and R3 also decreases
by 63.53\%, from 2.55\% under Base to 0.93\% under STAB.
Taken together, these results show that STAB strengthens bottleneck exposure
in 59 of the 60 aggregated comparisons while reducing sensitivity to the
selected evaluation solution on average.

\begin{table*}[h]
\centering
\small
\setlength{\tabcolsep}{3.4pt}
\renewcommand{\arraystretch}{1.08}
\caption{
ASR across the five evaluation solutions \(s_{p,r}\).
Fast and Slow correspond to \(s_{p,\mathrm{fast}}\) and
\(s_{p,\mathrm{slow}}\), respectively, and R1, R2, and R3 correspond to
\(s_{p,r_i}\) for \(i \in \{1,2,3\}\).
Range is the difference between the maximum and minimum ASR across the five
evaluation solutions.
}
\begin{tabular*}{\textwidth}{@{\extracolsep{\fill}}llccccccc@{}}
\toprule
\multicolumn{9}{c}{\textit{Python}} \\
\midrule
Model & Method & Fast & Slow & R1 & R2 & R3 & Range & Avg. \\
\midrule
\multirow{2}{*}{Qwen-3.5} 
& Base & 26.57\% & 57.10\% & 70.34\% & 59.90\% & 53.82\% & 43.77\%p & 53.55\% \\
& STAB & 56.14\% & 74.20\% & 75.85\% & 74.30\% & 73.53\% & 19.71\%p & \textbf{70.80\%} \\
\midrule
\multirow{2}{*}{Gemma-4} 
& Base & 42.32\% & 65.80\% & 43.29\% & 40.87\% & 41.16\% & 24.93\%p & 46.69\% \\
& STAB & 60.00\% & 79.23\% & 82.03\% & 79.61\% & 77.87\% & 22.03\%p & \textbf{75.75\%} \\
\midrule
\multirow{2}{*}{Gemini-3.1} 
& Base & 34.11\% & 54.88\% & 56.81\% & 54.88\% & 53.14\% & 22.70\%p & 50.76\% \\
& STAB & 59.23\% & 75.65\% & 79.81\% & 77.78\% & 75.94\% & 20.58\%p & \textbf{73.68\%} \\
\midrule
\multirow{2}{*}{GPT-5.4} 
& Base & 51.88\% & 68.21\% & 69.76\% & 68.50\% & 65.89\% & 17.88\%p & 64.85\% \\
& STAB & 56.04\% & 73.91\% & 76.81\% & 74.78\% & 72.85\% & 20.77\%p & \textbf{70.88\%} \\
\midrule
\multicolumn{9}{c}{\textit{Java}} \\
\midrule
Model & Method & Fast & Slow & R1 & R2 & R3 & Range & Avg. \\
\midrule
\multirow{2}{*}{Qwen-3.5} 
& Base & 27.60\% & 56.65\% & 69.05\% & 54.66\% & 54.57\% & 41.45\%p & 52.51\% \\
& STAB & 59.46\% & 75.11\% & 76.47\% & 77.83\% & 77.92\% & 18.46\%p & \textbf{73.36\%} \\
\midrule
\multirow{2}{*}{Gemma-4} 
& Base & 41.54\% & 57.38\% & 42.17\% & 41.45\% & 40.36\% & 17.02\%p & 44.58\% \\
& STAB & 63.08\% & 82.17\% & 82.35\% & 82.62\% & 83.98\% & 20.90\%p & \textbf{78.84\%} \\
\midrule
\multirow{2}{*}{Gemini-3.1} 
& Base & 35.48\% & 58.91\% & 60.54\% & 58.82\% & 60.00\% & 25.06\%p & 54.75\% \\
& STAB & 59.82\% & 77.65\% & 78.82\% & 80.00\% & 80.90\% & 21.08\%p & \textbf{75.44\%} \\
\midrule
\multirow{2}{*}{GPT-5.4} 
& Base & 57.47\% & 74.30\% & 73.30\% & 72.67\% & 75.48\% & 18.01\%p & 70.64\% \\
& STAB & 58.01\% & 75.66\% & 78.55\% & 76.65\% & 76.74\% & 20.54\%p & \textbf{73.12\%} \\
\midrule
\multicolumn{9}{c}{\textit{C++}} \\
\midrule
Model & Method & Fast & Slow & R1 & R2 & R3 & Range & Avg. \\
\midrule
\multirow{2}{*}{Qwen-3.5} 
& Base & 34.74\% & 57.19\% & 73.19\% & 60.81\% & 54.07\% & 38.45\%p & 56.00\% \\
& STAB & 59.33\% & 69.11\% & 71.70\% & 72.74\% & 72.37\% & 13.41\%p & \textbf{69.05\%} \\
\midrule
\multirow{2}{*}{Gemma-4} 
& Base & 58.07\% & 64.52\% & 41.33\% & 41.56\% & 40.59\% & 23.93\%p & 49.21\% \\
& STAB & 62.22\% & 71.78\% & 76.44\% & 77.48\% & 76.37\% & 15.26\%p & \textbf{72.86\%} \\
\midrule
\multirow{2}{*}{Gemini-3.1} 
& Base & 32.22\% & 42.52\% & 41.78\% & 44.59\% & 43.63\% & 12.37\%p & 40.95\% \\
& STAB & 58.52\% & 67.11\% & 72.74\% & 74.37\% & 74.67\% & 16.15\%p & \textbf{69.48\%} \\
\midrule
\multirow{2}{*}{GPT-5.4} 
& Base & 53.85\% & 61.63\% & 66.15\% & 66.74\% & 65.19\% & 12.89\%p & 62.71\% \\
& STAB & 60.30\% & 66.59\% & 71.19\% & 72.15\% & 72.15\% & 11.85\%p & \textbf{68.48\%} \\
\bottomrule
\end{tabular*}

\label{tab:solution_variance}
\end{table*}

\section{Full Adversarial Scenario Catalog} \label{app:scenario_table}
Tables~\ref{tab:full_catalog} and~\ref{tab:full_catalog_cont} report STAB's adversarial scenario catalog, covering 13 scenarios and 51 implementations.
For each implementation, the tables list its vulnerability class and the adversarial construction principle 
a test case must preserve.
The vulnerability class indicates whether the construction principle depends on input arrangement, 
exact values, or only the maximum admissible input size.
Three vulnerability classes are defined as follows.

\begin{itemize}[noitemsep,topsep=2pt]
    \item \emph{Structural.}
        A structural vulnerability arises when the relative arrangement of the input forces 
        an implementation toward its worst case.
        The exact values are not essential, provided that the arrangement remains unchanged.
        For example, quicksort with a first or last pivot runs in $O(n^2)$ time on any strictly ascending sequence, 
        since each partition removes only one element.
        Depth-first search via system recursion can overflow the call stack on a bamboo tree,
        since the recursion depth becomes $n$ on a chain of $n$ nodes.
        This class captures constructions whose effectiveness depends on shape, order, or adjacency rather than on exact numeric identities.
    \item \emph{Numerical.}
        A numerical vulnerability arises when the slow input must satisfy value-specific conditions 
        tied to the algorithm or to implementation parameters.
        The arrangement alone is insufficient, because replacing the values with arbitrary values of the same size 
        destroys the condition that triggers the slow path.
        For example, Euclidean GCD performs its maximum number of divisions on consecutive 
        Fibonacci values $(F_k, F_{k+1})$, because each quotient remains equal to one until the recurrence unwinds.
        An identity hash table with bucket count $B$ collapses into one bucket 
        when all keys satisfy $k_i \equiv 0 \pmod{B}$. Rabin--Karp with base $b$ and modulus $q$ can 
        degrade to repeated verification when the text and pattern are constructed so that distinct
        substrings share the same polynomial hash modulo $q$. 
        This class captures constructions whose effectiveness depends on arithmetic relations such as recurrence or divisibility.
    \item \emph{Size Only.}
        A size-only vulnerability arises when neither the arrangement nor the values
        change the implementation's asymptotic runtime.
        The slow input is obtained by setting the size parameters to the largest values
        allowed by the specification.
        For example, AVL tree operations keep height $O(\log n)$ for every insertion
        order, and KMP string matching scans the text and pattern in $O(n+m)$ time for
        all string pairs.
        This class captures implementations whose invariants make input volume the only
        effective way to increase runtime.
\end{itemize}

\begin{table*}[!t]
\small
\centering
\setlength{\tabcolsep}{4.0pt}
\renewcommand{\arraystretch}{1.10}
\caption{Full catalog of STAB's 13 adversarial scenarios across 51 implementations. Class labels are Structural, Numerical, and Size Only. The adversarial construction principle column summarizes trigger conditions rather than full generator recipes.}
\begin{tabularx}{\textwidth}{@{}p{1.75cm}p{4.35cm}p{1.65cm}>{\centering\arraybackslash}X@{}}
\toprule
Scenario & Implementation & Class & Adversarial Construction Principle\\
\midrule

\multirow{9}{1.75cm}{Sorting}
& Bubble sort                         & Structural & Strictly decreasing \(a_i=n-i\) \\
& Insertion sort                      & Structural & Strictly decreasing \(a_i=n-i\) \\
& Selection sort                      & Size Only  & Maximum \(n\), arbitrary permutation \\
& Merge sort                          & Structural & Alternating high and low values \\
& Quicksort~(first or last pivot)     & Structural & Monotone order against pivot position \\
& Quicksort~(median of three)         & Structural & Bentley--McIlroy sequence with extreme sampled median \\
& Quicksort~(random pivot)            & Structural & Exposed seed with extreme pivot ranks \\
& Heapsort~(Floyd's sift-down)        & Structural & Descending order causing long sift downs \\
& Shell sort                          & Structural & Gap-dependent interleaved descending order \\
\midrule

\multirow{6}{1.75cm}{Hashing}
& Hash table~(identity, int keys)     & Numerical  & Keys congruent modulo bucket count \\
& Hash table~(chain, no treeify)      & Numerical  & Equal hashCode strings such as Aa and BB \\
& Hash table~(chain and treeify)      & Numerical  & Equal hashCode strings with tree fallback \\
& Hash table~(SipHash, random seed)   & Numerical  & Exposed seed collision strings \\
& Rolling hash~(single modulus)       & Numerical  & Collisions under chosen base and modulus \\
& Rolling hash~(double modulus)       & Size Only  & Maximum volume with no feasible collision budget \\
\midrule

\multirow{5}{1.75cm}{Tree traversal}
& DFS via system recursion            & Structural & Bamboo chain of \(n\) vertices \\
& DFS via explicit stack              & Size Only  & Maximum \(n\), arbitrary shape \\
& Centroid decomposition              & Structural & Star tree centered at one vertex \\
& LCA~(naive parent walking)          & Structural & Bamboo tree with deepest leaf queries \\
& Tree DP~(single post-order)         & Size Only  & Maximum \(n\), arbitrary tree \\
\midrule

\multirow{6}{1.75cm}{Graph shortest path}
& Dijkstra~(lazy, PQ duplicates)      & Structural & Dense graph with repeated distance decreases \\
& Dijkstra~(proper heap)              & Size Only  & Maximum \(V\) and \(E\) \\
& Bellman--Ford                        & Size Only  & Maximum \(V\) and \(E\) \\
& Floyd--Warshall                      & Size Only  & Maximum \(V\) \\
& BFS~(unit weight)                   & Size Only  & Maximum \(V\) and \(E\) \\
& SPFA                                & Structural & Chain graph with strategic back edges \\
\bottomrule
\end{tabularx}
\label{tab:full_catalog}
\end{table*}

\begin{table*}[!t]
\small
\centering
\setlength{\tabcolsep}{4.0pt}
\renewcommand{\arraystretch}{1.10}
\caption{Full catalog of STAB's 13 adversarial scenarios across 51 implementations. Part 2 covers string matching, BST and segment structures, DP and recursion, number theory, binary search, two pointers, bitmasks, combinatorics, and geometry.}
\begin{tabularx}{\textwidth}{@{}p{1.75cm}p{4.35cm}p{1.65cm}>{\centering\arraybackslash}X@{}}
\toprule
Scenario & Implementation & Class & Adversarial Construction Principle \\
\midrule

\multirow{5}{1.75cm}{String matching}
& Naive string matching               & Structural & Text \(\mathrm{a}^{n-1}\mathrm{b}\), pattern \(\mathrm{a}^{m-1}\mathrm{b}\) \\
& Knuth--Morris--Pratt~(KMP)            & Size Only  & Maximum \(n\) and \(m\) \\
& Rabin--Karp~(single modulus)         & Numerical  & Colliding windows under chosen modulus \\
& Suffix array~(basic compare sort)   & Structural & Repetitive string with long common prefixes \\
& Z-algorithm                         & Size Only  & Maximum string length \\
\midrule

\multirow{5}{1.75cm}{BST / segment}
& BST (unbalanced)                    & Structural & Sorted or reverse-sorted insertion order \\
& AVL tree                            & Size Only  & Maximum \(n\) \\
& Splay tree                          & Size Only  & Maximum operations, amortized bound intact \\
& Treap~(randomized)                  & Structural & Exposed seed with monotone priorities \\
& Segment tree                        & Size Only  & Maximum \(n\) and \(q\) \\
\midrule

\multirow{3}{1.75cm}{DP / recursion}
& Unmemoized recursion                & Size Only  & Maximum \(n\) for exponential recurrence \\
& Memoized DP~(top-down)              & Size Only  & Maximum reachable state count \\
& Iterative bottom-up DP              & Size Only  & Maximum reachable state count \\
\midrule

\multirow{3}{1.75cm}{Number theory}
& Euclidean GCD                       & Numerical  & Consecutive Fibonacci pair \((F_k,F_{k+1})\) \\
& Sieve of Eratosthenes               & Size Only  & Maximum \(n\) \\
& Trial division                      & Numerical  & Prime or semiprime near \(N_{\max}\) \\
\midrule

\multirow{2}{1.75cm}{Binary search}
& Binary search on answer             & Structural & Full range with extreme boundary \\
& \(\mathrm{lower\_bound}\) on duplicates & Structural & All equal elements with repeated queries \\
\midrule

Two pointers
& Naive sliding window                & Structural & Inputs causing repeated full rescans \\
\midrule

\multirow{2}{1.75cm}{Bitmasks}
& Held--Karp / bitmask DP              & Structural & Maximum \(n\), all states reachable \\
& SOS DP~(naive submask iteration)    & Structural & Maximum \(n\), all entries nonzero \\
\midrule

\multirow{2}{1.75cm}{Combinatorics}
& Factorial without precompute        & Structural & Maximum \(n\), distinct full range queries \\
& Inclusion-exclusion                 & Structural & \(k=20\) to \(25\), all subsets nonzero \\
\midrule

\multirow{2}{1.75cm}{Geometry}
& Convex hull on collinear points     & Structural & Maximum collinear point set \\
& Closest pair on grid~(kd-tree)      & Structural & Integer grid with tied distances \\
\bottomrule
\end{tabularx}
\label{tab:full_catalog_cont}
\end{table*}

\end{document}